\documentclass[sigconf]{acmart}
\AtBeginDocument{%
  }
\usepackage{array}
\usepackage{url}
\usepackage{graphicx}
\usepackage{makecell}
\usepackage{amsmath}
\usepackage{float}
\usepackage{hyperref}

\copyrightyear{2025}
\acmYear{2025}
\setcopyright{acmlicensed}
\acmConference[AIQAM '25]{Proceedings of the 2nd
ACM Workshop in AI-powered Question \& Answering Systems}{October 27--28,
2025}{Dublin, Ireland}
\acmBooktitle{Proceedings of the 2nd ACM Workshop in AI-powered Question
\& Answering Systems (AIQAM '25), October 27--28, 2025, Dublin, Ireland}
\acmDOI{10.1145/3746274.3760394}
\acmISBN{979-8-4007-2056-7/2025/10}

\begin{document}
\title{Exploring the Application of Visual Question Answering (VQA) for Classroom Activity Monitoring}
\author{Sinh Trong Vu}
\email{sinhvt@hvnh.edu.vn}
\affiliation{%
  \institution{Banking Academy of Vietnam}
  \city{Hanoi}
  \country{Vietnam}
  }
\author{Hieu Trung Pham}
\email {hieupt05.work@gmail.com}
\affiliation{%
 \institution{Banking Academy of Vietnam}
  \city{Hanoi}
  \country{Vietnam}
  }
\author{Dung Manh Nguyen}
 \email {nguyenmanhdung0627@gmail.com}
\affiliation{%
 \institution{Banking Academy of Vietnam}
  \city{Hanoi}
  \country{Vietnam}
 }
\author{Hieu Minh Hoang}
\email {26a4041235@hvnh.edu.vn}
\affiliation{%
 \institution{Banking Academy of Vietnam}
  \city{Hanoi}
  \country{Vietnam}
  }  
    \author{Nhu Hoang Le}
\email{hoangnhule017@gmail.com}
\affiliation{%
  \institution{Banking Academy of Vietnam}
  \city{Hanoi}
  \country{Vietnam}
    }
\author{Thu Ha Pham}
\email {thucan041105@gmail.com}
\affiliation{%
 \institution{Banking Academy of Vietnam}
  \city{Hanoi}
  \country{Vietnam}
  }
\author {Tai Tan Mai}
\email{tai.tanmai@dcu.ie}
\affiliation{%
 \institution{Dublin City University}
  \city{Dublin}
  \country{Ireland}
}
\begin{abstract}
    Classroom behavior monitoring is a critical aspect of educational research, with significant implications for student engagement and learning outcomes. Recent advancements in Visual Question Answering (VQA) models offer promising tools for automatically analyzing complex classroom interactions from video recordings. In this paper, we investigate the applicability of several state-of-the-art open-source VQA models, including \textit{VideoLLaMA2, VideoLLaMA3, QWEN3, and NVILA}, in the context of classroom behavior analysis. To facilitate rigorous evaluation, we introduce our \textit{BAV-Classroom-VQA} dataset derived from real-world classroom video recordings at the Banking Academy of Vietnam. We present the methodology for data collection, annotation, and benchmark the performance of the selected VQA models on this dataset. Our initial experimental results demonstrate that all four models achieve promising performance levels in answering behavior-related visual questions, showcasing their potential in future classroom analytics and intervention systems.
\end{abstract}


\maketitle
\section{Introduction}
In the current educational environment, classroom activities generate a diverse and abundant range of data, which differs significantly from traditional classroom settings that primarily rely on textual \cite{Sinh2024aiqam} \cite{10.1145/2090116.2090132} or verbal information \cite{9157453}. Among these data types, visual data plays a pivotal role in reflecting the dynamics of teaching and learning interactions. Specifically, classroom images can offer rich visual cues about students’ emotional expressions, attention levels, and their engagement with both the lesson content and their peers \cite{9157453}. 

However, the analysis of this type of data mainly relies on manual observation and subjective interpretation, which can lead to bias and limit the scalability of its application \cite{10.1145/2090116.2090132}. To address this challenge, many recent works have sought to automate classroom observation and assessment through image-based analysis methods. Instead of relying solely on traditional indicators such as test scores or surveys, directly analyzing classroom activities through video can provide teachers with deeper insights into students' levels of attention and engagement \cite{automatevisual}.

Visual Question Answering has emerged as a prominent approach to this problem, allowing systems to generate answers to questions grounded in image or video data \cite{roboflowVQA2023}. Modern VQA methods have shifted from this joint encoding scheme to Vision Language Pre-training employing transformer architectures trained on generalized tasks with large image-text pair datasets and then fine-tuned to downstream tasks like VQA \cite{Ishmam_2024}, which opens up opportunities for applications in the educational domain. Integrating VQA into classroom monitoring has the potential to reduce human intervention while providing objective and real-time assessments of classroom status. This not only assists teachers in monitoring classroom activities more effectively but also generates valuable data to support the improvement of teaching methodologies. 

In this paper, we survey existing Visual Question Answering (VQA) models, including \textit{VideoLLaMA2, VideoLLaMA3, NVILA, and QWEN} and assess their potential for application in classroom video analysis. Specifically, the contributions of the paper are summarized as follows:
\begin{enumerate}
    \item Evaluate the potential of Visual Question Answering (VQA) in monitoring classroom behaviors and events, as well as in interpreting and explaining such behaviors through AI.
    \item Construct a specialized VQA dataset for classroom management, namely BAV-Classroom-VQA, including behavior annotations and diverse question-answer pairs that reflect real educational scenarios.
    \item Evaluate the quality and potential of state-of-the-art VQA models in the educational domain by conducting experiments on behavior recognition, event interpretation, and interactive querying in classroom videos.
\end{enumerate}

\begin{figure}
    \centering
    \includegraphics[width=1.1\linewidth]{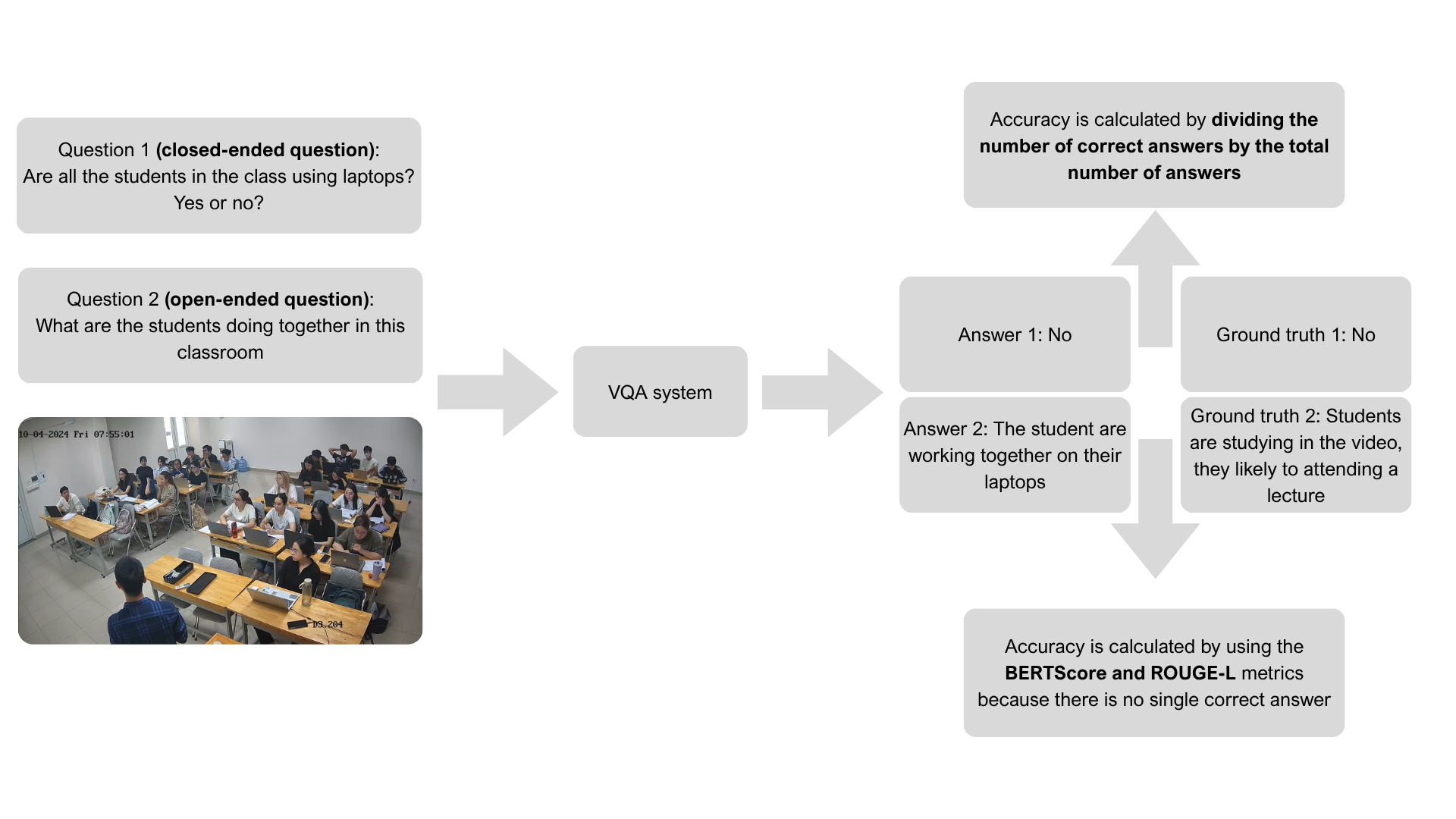}
    \caption{VQA Model Evaluation Framework}
    \Description{A diagram showing the VQA Model Evaluation Framework.}
    \label{fig:bang0}
\end{figure}

This work is conducted during the AIQAM'25, the second workshop in AI-powered Question and Answering Systems based on multimedia data\cite{aiqam25}.

\section{Related Work}
\subsection{Classroom Monitoring System with computer vision}
In recent years, numerous studies have integrated object detection and behavior classification techniques into classroom monitoring systems. Advanced models such as YOLOv5, YOLOv7 and their refined variants - YOLO-BRA \cite{yang2024studentclassroombehaviordetection}, YOLO-CBD \cite{peng2025classroombehavior}, or YOLOv8-EMA \cite {chen2023studentbehavior} have been employed in classroom and exam surveillance to identify student behavior: \textit{using mobile phone, bowing, raising hand} or \textit{taking notes}.These models have reported promising results, achieving mAP@0.5 ranging from 82\% to 88\% on domain-specific classroom datasets, including SCB-Dataset3 \cite {yang2024scbdataset3benchmark}, highlighting their effectiveness in recognizing student behaviors in crowded and complex classroom scenarios.\\
Despite recent advances, most existing approaches share a fundamental limitation: they process images or videos at the frame level, typically following a \textit{"detect–count–classify"} pipeline. As noted in \cite{Wang_2023}, they point out that although object detectors can localize and recognize all objects in a scene, they cannot understand the semantic relationships among these objects. This gap presents an opportunity for our research team to integrate Video Question Answering (VQA) into classroom monitoring systems, aiming not only to identify student behaviors but also to interpret and explain them through natural language queries (NLP). Such an approach paves the way for more advanced applications in analyzing educational interactions.
\subsection{Visual Question Answering systems}
Visual Question Answering (VQA) is  a key task in the field of Artificial Intelligence \cite{UPPAL2022149}, combining Computer Vision (CV) and Natural-Language Processing (NLP). In this task, the system is required to generate a natural language answer based on visual information from images or videos. In other words, VQA is like training the computer to not only "\textit{see}" the visual elements but also to "understand" and "\textit{speak}" about them when prompted with questions. 

Unlike traditional recognition tasks, QA requires the system to not only interpret visual content but also analyze the question and perform cross-modal reasoning to produce an appropriate answer. With the advancement of large language models, VQA has expanded from static image processing to Video Question Answering (VideoQA) - a task that requires understanding temporal dynamics and actions within video sequences.

At present, numerous advanced Visual Question Answering (VQA) models have been developed, with several widely adopted models demonstrating relatively high levels of accuracy in experimental settings. Prior research has indicated that, when compared to other open-source models, VideoLLaMA2 achieved the highest accuracy, attaining 51.7\% on the EgoSchema (MC-VQA) dataset and 71.7\% on the MSVU (OE-VQA) dataset \cite{cheng2024videollama2}. VideoLLaMA3 exhibited superior accuracy relative to other open-source models when evaluated across various datasets such as ActivityNet-QA, NextQA, and LVBench \cite{zhang2025videollama3}.  In terms of both performance and accuracy, NVILA has also received high evaluations, achieving outstanding accuracy on datasets including NeXT-QA (82.2\%), MLVU (70.1\%), and ActivityNet-QA (60.9\%) \cite{liu2025nvilaefficient}. Similarly, Qwen3 has demonstrated performance surpassing that of strong models such as DeepSeek-R1 and DeepSeek-V3, and moreover, has shown competitiveness with several leading closed-source models \cite{yang2025}.

Building appropriate datasets is a key factor in the development of VQA models. Currently, popular VQA datasets such as VQA v2 and GQA mainly focus on static images \cite{ma2024}, \cite{Sirignano_2018}], while datasets like TGIF-QA and MSRVTT-QA are designed for VideoQA, targeting questions about actions, events, and temporal order \cite{jang2017}, \cite{jang2017tgifqa}.Some others, such as HowToVQA69M and ActivityNet-QA, extend the scope to instructional or action videos \cite{yang2021}, \cite{yu2019}.

From the aforementioned findings, it is clear that VQA and VideoQA have made significant technological and application advances. However, none of the existing datasets reflect the specific context of \textit{classroom environments}, which involves complex learning behaviors and interactions between teachers and students. These scenarios present unique challenges that require dedicated datasets and models tailored for classroom analysis. For this reason, our research team proposes to explore the application of VQA models in classroom video analysis.  As a first step, we conduct experiments on real classroom videos to evaluate the feasibility of this approach.

\subsection{Question types in VQA datasets}
According to the classification in \cite{zhong2022}, questions in VideoQA tasks are typically divided into two main categories: Factoid and Inference. \textit{Factoid questions} focus on directly recognizing visual information in the video, such as objects, locations, or specific actions. This category usually deals with questions that require identifying explicit and observable elements within the video content. In contrast, \textit{Inference questions} require the model to reason over the events depicted in the video, integrating background knowledge or analyzing causal relationships and temporal sequences to generate appropriate answers. Additionally, this article also highlights several common subtypes of questions in VideoQA tasks, such as counting objects, recognizing specific features of objects, identifying spatial relations, or reasoning about cause-effect and temporal order. 

Beyond question content, we also refer to the approach in \cite{huynh2025visualquestionansweringearly}, where VQA questions are classified based on the answer format. These include \textbf{\textit{Close-Ended questions}}, such as \textit{Yes/No} or \textit{multiple-choice}, and \textbf{\textit{ Open-Ended questions}}, where the model is expected to produce free-form answers that may involve detailed descriptions or explanations depending on how the question is posed.

\section{Dataset construction and Research design}
\subsection{Classroom video collections}
To construct a dataset for Classroom Behavior Visual Question Answering (VQA), we requested permission to access the classroom video repository at the International School – Banking Academy. From this repository, we intend to extract specific videos that have been agreed upon by both lecturers and students for research purposes. Specifically, we collected \textit{13 raw recording videos} with a total duration of \textit{11 hours and 58 minutes}, capturing classroom activities in different rooms.

Classroom recording videos were selected based on the following criterias:
 \begin{itemize}
     \item High image resolution, clear lighting, no blur or visual artifacts, and a wide-angle view that captures the full classroom context.
     \item All relevant subjects (lecturers and students) must be clearly visible within the frame.
     \item A variety of distinct and observable behaviors are present to facilitate question formulation.
 \end{itemize}
After selection, the videos were segmented into shorter clips (\textit{20 to 30 seconds}) to retain key information, reduce the data volume for model processing, and minimize the risk of behavior mislabeling by annotators.

To ensure privacy protection, all personally identifiable information such as faces, names, or any other personal markers not essential to the research was blurred or masked. All recordings were stored on secure, access-controlled devices and were never shared through public platforms. Access to the data was strictly limited to authorized research team members.

\subsection{Dataset construction}
As mentioned earlier, Video Question Answering (VQA) models have been extensively studied and applied across various domains to answer questions based on visual content from videos. However, after reviewing related studies, we found that no open dataset specifically designed for classroom monitoring with VQA exists. Therefore, we draw upon several studies on Visual Question Answering (VQA) to classify the questions before constructing the dataset.

Then, we developed a set of questions for classroom monitoring and management, based on the short video clips previously extracted. The questions are designed to assess various cognitive aspects and reasoning abilities of the models, covering both Close-ended and Open-ended formats.
The question types include Single-Choice, Multiple-Choice, Factual, Reasoning, Spatial. The structure and formulation of these questions were adapted from existing VQA datasets to ensure alignment with common benchmarks. Furthermore, based on certain contextual considerations within the classroom setting, we excluded several types of questions – namely Attribute Classification, Comparative, and Multiple-Choice – as also observed in \cite{huynh2025visualquestionansweringearly}. The structure and formulation of these questions were adapted from existing VQA datasets to ensure alignment with common benchmarks.

\begin{figure}
    \centering
    \includegraphics[width=1.1\linewidth]{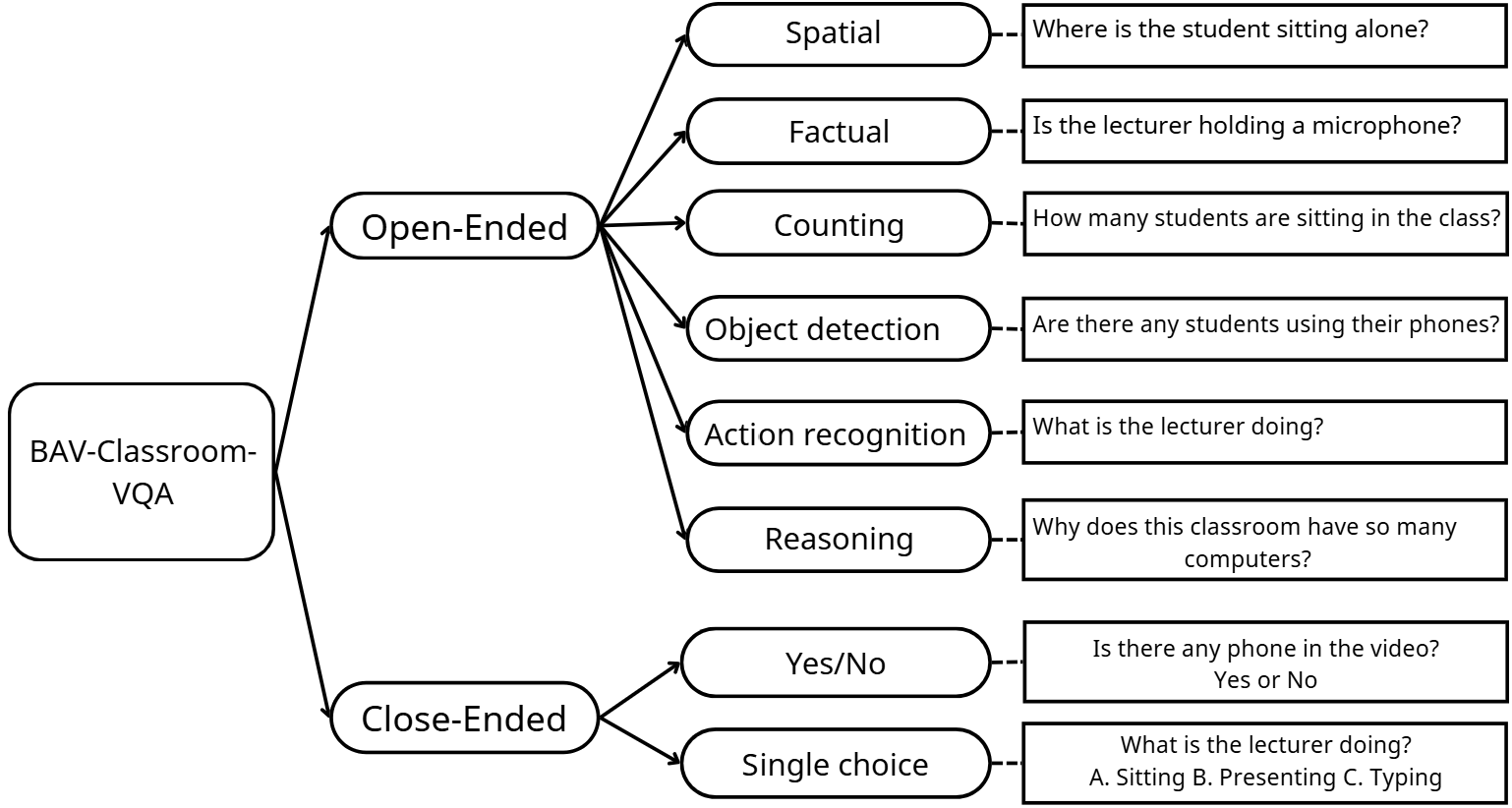}
    \caption{BAV-Classroom-VQA Dataset Structure}
    \Description{Structure of the BAV-Classroom-VQA dataset as shown in the figure.}
    \label{fig:dts}
\end{figure}

 In this setup, Close-ended questions are designed with predefined answer options, allowing the model to provide concise responses. In contrast, Open-ended questions are crafted to elicit reasoning or explanations. However, to maintain control over the expected answer length and ensure compatibility with model evaluation, we introduce constraints such as word limits and guided prompts such as “Not more than 3 words,” “Only one word or number,” or “Short sentence.” After constructing the question set, our annotators conduct a cross-review process of the answers and discuss the outcomes to produce an objective and accurate dataset.
 
 \subsection{Research Design}
 \subsubsection{Evaluated VQA Models}

 To evaluate the capabilities and potential of various VQA models within educational systems, our research team conducted experiments on 4 selected open-source VQA models, which were chosen based on the prior studies discussed in the preceding section. Specifically, these models include LLaMA2.1-7B-16F, LLaMA3-7B, NVILA-8B, and Qwen3-235B-A22B. These are widely used and highly regarded among open-source VQA models. 
The videos and corresponding questions were fed into the four selected models for evaluation, in which VideoLLaMA2, VideoLLaMA3: downloaded from Hugging Face \footnote{\url{https://huggingface.co/spaces/lixin4ever/VideoLLaMA2}}
, while Qwen3, NVILA: tested via the official websites of each model \footnote{\url{https://chat.qwen.ai/}} \footnote{\url{https://vila.hanlab.ai/}}. 

For the hardware setup, a computer equipped with a 13th Gen Intel Core i5-13420H processor (12 cores, ~2.1 GHz), 16 GB of RAM, and an NVIDIA GeForce RTX 2050 graphics card was used for the experiments.

\subsubsection{Evaluation Metrics}
To ensure objective and reliable evaluation of model performance and accuracy on both datasets and question types, our study adopts distinct assessment criteria for Closed-Ended and Open-Ended questions.

For Closed-Ended questions, where responses are selected from a predefined set of options, evaluation is relatively straightforward. Following the methodology from prior research \cite{huynh2025visualquestionansweringearly}, the \textbf{accuracy} is calculated as the proportion of correctly answered questions over the total number of questions:
\textit{\[
\text{Accuracy} = \frac{\text{Number of correct answers}}{\text{Total number of questions}}
\]
}

In contrast, evaluating Open-Ended questions is more complex due to the varied ways an answer can be expressed. The \textbf{ROUGE-L} metric \cite{lin-2004-rouge}, which measures similarity based on the longest common subsequence (LCS), is useful for assessing surface-level structural alignment but fails to capture semantically equivalent but differently phrased responses. On the other hand, \textbf{BERTScore} \cite{10964149}, which leverages contextual embeddings from BERT to compute semantic similarity between text pairs, offers a more robust understanding of meaning but is computationally intensive due to its reliance on large embedding models.

Recognizing the distinct advantages and limitations of each method, this study calculates and reports scores using both BERTScore and ROUGE-L to evaluate the consistency of structure and information order in the models’ responses.

\section{Results and Analysis}
After the experimental process and evaluation, the research team compiled statistical results to provide an overall assessment of the models.
\subsection{Close-Ended Results}
\subsubsection{Result Overview}
\begin{table}[H]
    \centering
    \begin{tabular}{|m{2.3cm}|
                 >{\centering\arraybackslash}m{1.1cm}|
                 >{\centering\arraybackslash}m{1.1cm}|
                 >{\centering\arraybackslash}m{1.1cm}|
                 >{\centering\arraybackslash}m{1.1cm}|
                 }
    \hline
     \textbf{Model}  & \textbf{LLaMA2}  &\textbf{LLaMA3 }& \textbf{QWEN3}  & \textbf{NVILA} \\ 
    \hline
    Accuracy Yes/No & \textbf{80\%} & 75\% & \textbf{80\%} & 50\% \\ 
    \hline
    Accuracy Single Choice & \textbf{61.9\%} & 57.1\% & 47.6\% & 57.1\% \\
    \hline
    AVG response time & 20s & $\sim$10s & \textbf{5-10s} & 30-60s \\
    \hline
    \end{tabular}
    \caption{VQA Model Performance(Close-Ended Questions).}
    \label{tab:bang1}
\end{table}
From the table above, it can be observed that the models performed better with \textbf{\textit{Yes/No}} questions. Notably, both VideoLLaMA2 and QWEN3 achieved 80\% accuracy, demonstrating their strong capabilities in handling simple binary tasks.
In contrast, \textbf{\textit{Single Choice}} questions yielded lower accuracy, ranging from 47\% to 61\%, with QWEN3 being the lowest performer. Regarding \textit{\textbf{response time}}, QWEN3 was the fastest (5–10 seconds), followed by VideoLLaMA3 (~10 seconds), VideoLLaMA2 (20 seconds), and NVILA (30–60 seconds).
\subsubsection{Model Evaluation}
For this type of close-ended question, VideoLLaMA2 demonstrated the best overall performance, achieving 80\% accuracy on Yes/No questions and 61.9\% on Single Choice questions. This model excels at tasks such as behavior identification, basic object detection, and simple counting. However, it still struggles with questions requiring additional information such as location, color, or recognizing multiple simultaneous actions. Similarly, VideoLLaMA3 performs well with simple object identification and counting tasks. Although its accuracy for close-ended questions is slightly lower than VideoLLaMA2, the difference is not significant. Additionally, VideoLLaMA3 responds twice as fast as its predecessor.
QWEN3 achieved strong performance in Yes/No questions (80\%) with the fastest response time (5–10 seconds). However, it had the lowest accuracy in Single Choice questions (47.6\%), suggesting that this model is suitable for immediate binary responses but less stable for complex multiple-choice tasks. Finally, NVILA showed unimpressive performance, scoring 50\% for Yes/No questions and 57.1\% for Single Choice questions. The model encountered difficulties with spatial reasoning, group behavior analysis, and detecting occluded objects, indicating that it is not yet suitable for multiple-choice tasks requiring both high accuracy and rapid responses.
\subsection{Open-Ended Questions Results}
We also tested the models on Open-Ended questions and provided detailed evaluations for each.

\begin{table}[H]
    \centering
    \begin{tabular}{|m{1.3cm}|
                 >{\centering\arraybackslash}m{1.5cm}|
                 >{\centering\arraybackslash}m{1.5cm}|}
    \hline
     \textbf{Model}  & \textbf{BERT Score}  & \textbf{ROUGE-L} \\
    \hline
    \textbf{NVILA}  & \textbf{0.7462} & \textbf{0.4134 } \\ 
    \hline
    \textbf{LLaMA3} & 0.7120 & 0.3593  \\ 
    \hline
    \textbf{LLaMA2} & 0.6877 & 0.3075  \\ 
    \hline
    \textbf{QWEN3} & 0.7353 & 0.4013  \\
    \hline
    \end{tabular}
    \caption{VQA Model Performance(Open-Ended Questions)}
    \label{tab:bang_4}
\end{table}

\subsubsection{Result Overview}
Overall, NVILA and QWEN3 achieved the highest scores in both semantic similarity (BERTScore) and structural matching (ROUGE-L). In contrast, the two LLaMA models (VideoLLaMA2 and VideoLLaMA3) performed lower on both metrics. 

\begin{table}[H]
    \centering
    \begin{tabular}{|m{1.45cm}|
                 >{\centering\arraybackslash}m{0.85cm}|
                 >{\centering\arraybackslash}m{0.85cm}|
                 >{\centering\arraybackslash}m{0.8cm}|
                 >{\centering\arraybackslash}m{0.9cm}|
                 >{\centering\arraybackslash}m{0.8cm}|
                 >{\centering\arraybackslash}m{0.8cm}|
                 }
    \hline
    \textbf{Type}  & \textbf{Spatial}  & \textbf{Factual} & \textbf{Coun- ting}  & \textbf{Object Detec.} & \textbf{Action Recog.} & \textbf{Reaso- ning} \\
    \hline
    \textbf{NVILA}
    \textit{(BERT Score)} & \textbf{0.7235} & \textbf{0.9527} & 0.8670 & 0.8394 & \textbf{0.5838} & 0.5201 \\
    \hline
    \textbf{NVILA}
    \textit{(ROUGE-L)}& \textbf{\textit{0.4397}} & \textbf{\textit{0.8000}} & 0.2000 & \textbf{\textit{0.5308}} & \textbf{\textit{0.3372}} & 0.1726 \\
    \hline
    \textbf{LLaMA3}
    \textit{(BERT Score)}& 0.6698 & 0.9290 & 0.7265 & \textbf{0.9287} & 0.5005 & 0.5182 \\
    \hline
    \textbf{LLaMA3}
    \textit{(ROUGE-L)}& \textbf{\textit{0.2668}} & \textbf{\textit{0.7000}} & 0.2000 & \textbf{\textit{0.7000}} & 0.2236 & \textbf{\textit{0.2653}} \\
    \hline
    \textbf{LLaMA2}
    \textit{(BERT Score)}& 0.6535 & 0.9053 & 0.7608 & 0.8093& 0.5501 & 0.4473 \\
    \hline
    \textbf{LLaMA2}
    \textit{(ROUGE-L)}& 0.2417 & 0.6000 & 0.2000 & 0.4000 & 0.2530 & 0.1500 \\
    \hline
    \textbf{QWEN3}
    \textit{(BERT Score)}& 0.6462 & 0.9290& \textbf{0.8713} & 0.8721& 0.5367 & \textbf{0.5405} \\
    \hline
    \textbf{QWEN3}
    \textit{(ROUGE-L)}& 0.3581 & 0.7000 & 0.2000 & 0.6429 & 0.2657 & \textbf{\textit{0.2245}} \\
    \hline
    \end{tabular}
    \caption{BERT score and ROUGE-L score by question type for VQA Models}
    \label{tab:bang_2}
\end{table}

When breaking down the Open-Ended questions by category, the Factual and Object Detection questions yielded the highest scores, ranging from \textit{0.7 }to \textit{0.9}. These questions primarily focus on object presence and recognition, which are relatively straightforward tasks. The Spatial and Counting questions resulted in slightly lower scores (\textit{0.5–0.6}), due to the additional complexity of spatial relationships and the presence of distracting factors in the video. Finally, the Action Recognition and Reasoning tasks, which require higher levels of inference and contextual understanding, scored the lowest—around 0.4—highlighting a current weakness across all models.

\subsubsection{Model Evaluation}
Among the evaluated models, \textbf{NVILA} achieved the highest performance in \textbf{Spatia}l (BERT Score: 0.7235, ROUGE-L: 0.4397), \textbf{Factual} (0.9527, 0.8000), and \textbf{Action Recognition} (0.5838, 0.3372) question types. This indicates its strength in information extraction, image structure analysis, and basic behavior recognition. However, for \textbf{Object Detection}, \textbf{Counting}, and \textbf{Reasoning questions}, NVILA demonstrated lower performance compared to certain other models. 

Furthermore, if combined with slower response times, it may be more suitable for analytical tasks that do not require real-time processing. QWEN3 exhibited stable performance without recording the lowest scores in any question type and showed particularly strong results in Factual (0.9290, 0.7000) and Object Detection (0.8721, 0.6429). It also outperformed other models in Counting (0.8713, 0.2000) and Reasoning (0.5405, 0.2245). Therefore, QWEN3 is suitable for diverse tasks requiring robust and generalizable processing capabilities. 

In contrast, VideoLLaMA3 achieved the highest results in Object Detection (0.9287, 0.7000) but showed lower performance in Counting (0.7265, 0.2000) compared to other models. Unlike the others, this model primarily generates counting answers in textual form. This indicates that its performance is lower in tasks requiring precise counting or object detection. 

Finally, VideoLLaMA2 recorded the lowest scores across most question types, particularly in Factual and Reasoning, reflecting its limited capability in handling open-ended questions that require higher-order reasoning. We extracted some examples from our experiments demonstrating the differences in outputs of each model in Table \ref{tab:bang_3}.

\begin{table}[H]
    \centering
    \renewcommand{\arraystretch}{1.25}
    \begin{tabular}{|m{1.1cm}|
                 >{\centering\arraybackslash}m{1.95cm}|
                 >{\centering\arraybackslash}m{2.8cm}|
                 >{\centering\arraybackslash}m{1.85cm}|
                 }
    \hline
    \textbf{Question (Category)}  & How many students are sitting in the class? Only one word or number. \textbf{(Counting)} & Where is the student wearing a mask sitting? Short sentence. \textbf{(Spatial)} & Is the lecturer holding a microphone? Only one word or number. \textbf{(Factual)} \\
    \hline
     \textbf{Ground Truth Answer} & 24 & The fourth table from the front in the rightmost row & Yes \\
    \hline
   \textbf{NVILA Answer} & 30 & The student wearing a mask is sitting in the back row, on the right side of the classroom. & Yes \\   
    \hline
    \textbf{LLaMA3 Answer} & 20 & At the back of the classroom. & Yes \\
    \hline
    \textbf{LLaMA2 Answer} & Many & Desk & Yes \\
    \hline
    \textbf{QWEN3 Answer} & 12 & The student wearing a mask is sitting in the second row from the front, on the left side of the classroom. & Yes \\
    \hline
    \textbf{\textit{Result}} & \textit{All incorrect} & \textit{NVILA describes most exactly} & \textit{All correct} \\
    \hline
    \end{tabular}
    \caption{Sample Open-Ended questions and answers from each model}
    \label{tab:bang_3}
\end{table}

\section{Conclusion and Future Works}
\subsection{Conclusion}
In this paper, we conducted a comprehensive survey on the potential application of Visual Question Answering (VQA) systems in monitoring classroom activities from video data. The BAV-Classroom-VQA dataset is collected from real-world classroom video recordings at the Banking Academy of Vietnam and annotated carefully. 

Our experimental results demonstrate that current VQA models, such as QWEN-VL, NVILA, and LLaMA-based approaches, can handle Close-Ended questions effectively, particularly in Yes/No and Single Choice tasks. However, when dealing with Open-Ended questions, especially those requiring reasoning or action recognition, the models still exhibit significant limitations. Despite these challenges, our study highlights the feasibility of applying VQA to real classroom scenarios, especially for quick recognition tasks like counting students or detecting object usage.

This approach presents practical potential to assist teachers in semi-automated classroom monitoring, offering real-time insights into student behaviors, attention levels, and interactions during lectures. The ability to analyze such aspects from video data can contribute to improving the overall quality of teaching and learning management.
\subsection{Future Works}
For future research, several directions are proposed to improve and extend the application of VQA systems in classroom monitoring.

First, data expansion is essential, involving the collection of a more diverse set of classroom videos with various environments, camera angles, and types of student interactions. This will help the models capture richer features and enhance their generalization capabilities. In addition, the research team will add test samples in future versions of the dataset to enable more thorough evaluation. Second, model improvement should be considered by incorporating domain-specific fine-tuning strategies, such as instruction tuning or domain adaptation tailored to educational data. 

Additionally, exploring newer multimodal models like GPT-4o or upcoming Visual Language Models (VLMs) released in 2025 could further improve performance. Another important direction is the development of real-time VQA systems, allowing teachers to receive immediate assistance during lectures. Finally, a comprehensive evaluation framework is necessary, extending beyond accuracy to include factors such as latency, robustness across different classroom settings, and the long-term impact of real-world deployment.

\section{Acknowledgments}

This research was supported by funding from the Banking Academy of Vietnam. The authors gratefully acknowledge the financial assistance that made this study possible.

\bibliographystyle{IEEEtran}     
\bibliography{references}
\end{document}